\documentclass[runningheads]{llncs}
\usepackage[T1]{fontenc}
\usepackage{graphicx}
\usepackage{amsmath}
\usepackage{amssymb}
\usepackage{hyperref}
\usepackage{physics} % partial derivative stuffs
\usepackage{bibentry}
\usepackage{algorithm}
\usepackage{soul}
\usepackage{authblk}
\usepackage{blindtext}
\usepackage{tikz}
\usepackage[noend]{algpseudocode}
\begin{document}
\title{An Algebraic Framework for Hierarchical Probabilistic Abstraction}
\author{Nijesh Upreti, Vaishak Belle}
\authorrunning{N. Upreti and V. Belle}
\institute{The University of Edinburgh, 10 Crichton Street, Edinburgh EH8 9AB, UK}
\maketitle              % typeset the header of the contribution
\begin{abstract}
Abstraction is essential for reducing the complexity of systems across diverse fields, yet designing effective abstraction methodology for probabilistic models is inherently challenging due to stochastic behaviors and uncertainties. Current approaches often distill detailed probabilistic data into higher-level summaries to support tractable and interpretable analyses, though they typically struggle to fully represent the relational and probabilistic hierarchies through single-layered abstractions. We introduce a hierarchical probabilistic abstraction framework aimed at addressing these challenges by extending a measure-theoretic foundation for hierarchical abstraction. The framework enables modular problem-solving via layered mappings, facilitating both detailed layer-specific analysis and a cohesive system-wide understanding. This approach bridges high-level conceptualization with low-level perceptual data, enhancing interpretability and allowing layered analysis. Our framework provides a robust foundation for abstraction analysis across AI subfields, particularly in aligning System 1 and System 2 thinking, thereby supporting the development of diverse abstraction methodologies.

\end{abstract}
\section{Introduction}
Abstraction serves as a core concept across knowledge domains, simplifying complex relationships into more comprehensible forms by focusing on essential details and discarding irrelevant ones \cite{giunchiglia1992theory,saitta2013abstraction,Abstraction_Belle}. In probabilistic models, abstraction is challenging due to inherent uncertainties and the stochastic nature of the systems involved \cite{ludtke2018state,reasoning-about-uncertainty}. Various methodologies have been developed to summarize detailed probabilistic information into higher-level representations while preserving essential characteristics and relationships, making such frameworks tractable and interpretable under uncertainty \cite{Abstraction_Belle,holtzen2018sound,holtzen2017probabilistic,probabilistic-graphical-models:-principles,machine-learning:-a-probabilistic-perspective}. Yet, single-layered probabilistic abstractions are often inadequate for capturing the full relational and probabilistic hierarchy in a single low-to-high-level mapping.

Recent works emphasize multi-layered representations to model real-world hierarchical complexities, such as in disease diagnosis \cite{disease_manage_h} and epidemic modeling \cite{H_epidemic_model}. These layered approaches are essential for analyzing phenomena across multiple levels, providing interpretable insights into complex interactions and uncertainties \cite{reasoning-about-knowledge-and-probability,reasoning-about-uncertainty,halpern2004representation}. Further, hierarchical modeling using directed acyclic graphs (DAGs) has become instrumental in fields like cognitive science, where hierarchical DAGs model complex concepts across layers \cite{danks2014unifying}. However, constructing interpretable high-level theories from low-level data requires a structured framework for hierarchical relational and probabilistic information. A formal mathematical treatment of abstraction hierarchies is thus critical, offering a robust foundation for reasoning in complex spaces. Starting from single-layer mappings and progressing to multi-layered models captures complexities at each level while clarifying relationships between different layers \cite{halpern2004representation}. Such a framework would further enable the development of methods to learn abstraction hierarchies in diverse contexts, supporting detailed decomposition for in-depth analysis \cite{PAH_Koller,HierarchialProb_Abs_Refinement}.

To position our work within this existing research landscape, we draw upon several foundational studies. First, Hofmann and Belle's work established an abstraction framework for robot programs with probabilistic beliefs, utilizing the DS logic and bisimulation in probabilistic dynamic systems to demonstrate sound and complete abstraction \cite{Hofmann_Belle}. Second, Beckers et al. contributed a framework for analyzing abstraction and approximation within causal models, providing formal definitions and results for approximate causal abstractions \cite{beckers2020_UAI}. Third, Segal et al.'s study on probabilistic abstraction hierarchies (PAH) introduced a principled approach for learning abstraction hierarchies from data, addressing complexities in modeling multi-level systems \cite{PAH_Koller}. Our work is inspired by these research endeavors and directly builds on Holtzen et al.’s papers on probabilistic program abstraction, which establishes single-layered mappings from concrete programs to their abstract versions with soundness properties \cite{holtzen2018sound,holtzen2017probabilistic}.

In this paper, we propose a hierarchical probabilistic abstraction framework that generalizes the measure-theoretic approach as introduced by Holtzen et al. to multi-layered settings while providing a structured, modular foundation for abstraction in complex systems. By decomposing abstraction into layered mappings, the framework supports detailed, layer-specific analysis alongside a comprehensive system-level understanding, enabling dual-level reasoning: individual transformations and probabilistic relationships at each layer, and cumulative effects across layers. This structured approach facilitates exploration of abstractions across AI subfields, notably in System 1 and System 2 cognition, by connecting high-level conceptual reasoning with low-level perceptual data \cite{Kahneman2011-KAHTFA-2}. This comprehensive perspective is vital for advancing research in cognitive AI, statistical relational learning, and neurosymbolic AI, laying a solid foundation for the development of diverse abstraction methodologies.

\section{Motivating Example}

Imagine a personalized learning system designed to optimize educational outcomes for students with diverse backgrounds and abilities. The goal is to create an adaptive education platform that tailors learning experiences based on individual cognitive profiles, learning environments, and behavioral patterns. This example is a slightly more complex version of examples as outlined in Belle's work \cite{Abstraction_Belle}. Towards modeling such a system, consider a probabilistic relational model (PRM) for a university database \(U\), representing student learning and performance. The model instantiates constraints for a parameterized Bayesian network as follows:
\[
\scriptsize
\text{Environmental Factors} \rightarrow \text{Cognitive Processes} \leftarrow \text{Learning Behaviors} \rightarrow \text{Educational Outcomes}
\]

\begin{multicols}{2}

\noindent
\textbf{Low-Level Theory \(U_l\):}
\begin{itemize}
\scriptsize
    \item[$\square$] $\textbf{0.7} \quad \text{HLE}(x, \text{High})$
    \item[$\square$] $\textbf{0.2} \quad \text{HLE}(x, \text{Medium})$
    \item[$\square$] $\textbf{0.1} \quad \text{HLE}(x, \text{Low})$
    \item[$\square$] $\textbf{0.6} \quad \text{PI}(x, \text{High})$
    \item[$\square$] $\textbf{0.3} \quad \text{PI}(x, \text{Medium})$
    \item[$\square$] $\textbf{0.1} \quad \text{PI}(x, \text{Low})$
    \item[$\square$] $\textbf{0.5} \quad \text{SES}(x, \text{High})$
    \item[$\square$] $\textbf{0.3} \quad \text{SES}(x, \text{Medium})$
    \item[$\square$] $\textbf{0.2} \quad \text{SES}(x, \text{Low})$
    \item[$\square$] $\textbf{0.4} \quad \text{WM}(x, \text{High}) \land \text{HLE}(y, \text{High}) \land \text{PI}(z, \text{High}) \land \text{SES}(w, \text{High}) \\ 
    \implies \text{cogAbility}(x, y, z, w, u)$ for $u \in \{\text{Strong}, \text{Moderate}\}$
    \item[$\square$] $\textbf{0.6} \quad \text{WM}(x, \text{Medium}) \land \text{HLE}(y, \text{Medium}) \land \text{PI}(z, \text{Medium}) \land \text{SES}(w, \text{Medium}) \\
    \implies \text{cogAbility}(x, y, z, w, u)$ for $u \in \{\text{Moderate}, \text{Weak}\}$
\end{itemize}

\columnbreak

\noindent
\textbf{High-Level Theory \(U_h\):}
\begin{itemize}
\scriptsize
    \item[$\square$] $\textbf{0.7} \quad \text{engagement}(x, \text{High})$
    \item[$\square$] $\textbf{0.3} \quad \text{engagement}(x, \text{Low})$
    \item[$\square$] $\textbf{0.5} \quad \text{cogAbility}(x, \text{Strong}) \land \\ \; \text{engagement}(y, \text{High}) \land \text{adaptation}(z, \text{High}) \\ \implies \text{academicPerformance}(x, y, z, u)$ for $u \in \{\text{Excellent}, \text{Good}\}$
    \item[$\square$] $\textbf{0.5} \quad \text{cogAbility}(x, \text{Weak}) \land \\ \text{engagement}(y, \text{Low}) \land \text{adaptation}(z, \text{Low}) \\ \implies \text{academicPerformance}(x, y, z, u)$ for $u \in \{\text{Average}, \text{Poor}\}$
\end{itemize}

\end{multicols}
In the given example case, while a two-layer abstraction from \(U_l\) to \(U_h\) captures some relationships between variables, it falls short due to the complexity and interdependencies in educational systems. In the formal model, we see numerous variables such as Home Learning Environment (HLE), Parental Involvement (PI), and Socioeconomic Status (SES), each affecting Cognitive Abilities (WM) and Learning Styles (LS). The interaction of these variables in a two-layer mapping is insufficient to capture the nuanced impacts across all layers. For instance, a high-quality home learning environment (HLE = High) might significantly enhance working memory (WM = High), which in turn influences engagement (engagement = High) and ultimately leads to better academic performance (academicPerformance = Excellent). However, this pathway also intersects with other variables like parental involvement and socioeconomic status, creating a web of interdependencies that a two-layer model cannot fully address.

To address complex interdependencies, we can use hierarchical abstractions to decompose the problem into multiple layers, each capturing specific relationships and interactions. This approach accommodates detailed representations that reflect the complex interdependencies among variables, such as those between home learning environment, parental involvement, and socioeconomic status. By maintaining probabilistic integrity across layers, this hierarchical model breaks down these complexities, providing interpretable insights that can guide educators and policymakers in addressing the nuanced factors impacting educational outcomes.

\subsection{New Directions in Hierarchical Probabilistic Abstraction}
In exploring probabilistic abstraction, we encounter pivotal questions that challenge our understanding of hierarchical models. These largely unexplored questions require a structured methodology to articulate the foundational principles of hierarchical abstraction. Our approach—constructing complex abstraction layers from foundational taxonomy—serves as a starting point to address these inquiries, offering a systematic way to navigate and clarify the complexities of such models.

Before delving into a detailed examination of hierarchical abstractions, it is essential to outline the key questions guiding our inquiry:
\begin{enumerate}
  \item \textit{Defining Hierarchical Abstraction:} What defines a hierarchical abstraction within probabilistic systems, and how can this definition be systematically applied across layers?
  \item \textit{Validating Hierarchical Structures:} How do we ascertain the accuracy of a hierarchical model in abstracting its underlying probabilistic layers?
  \item \textit{Navigating Abstraction Layers:} What mechanisms facilitate the effective translation and navigation between different levels of abstraction within a hierarchical model?
  \item \textit{Managing Complexity:} How does hierarchical abstraction aid in the simplification and management of the inherent complexity in probabilistic systems?
  \item \textit{Operationalizing Hierarchical Compositions:} How can operations and compositions across various hierarchical layers be conceptualized and applied to enhance probabilistic modeling?
\end{enumerate}

\section{Foundational Taxonomy for Probabilistic Abstraction}
In this section \footnote{For any schematic representation presented in this paper, the upward arrow ($\uparrow$) symbolizes an abstraction operation, indicating the transformation from a lower-level probability space to a higher-level abstracted probability space. Each step in the chain, represented by a unique abstraction operation $\mathcal{A}_i$, transitions the system from one probabilistic space $(\Omega_{i-1}, \Sigma_{i-1}, P_{i-1})$ to another $(\Omega_i, \Sigma_i, P_i)$.}, we delineate the taxonomy of abstraction processes by examining one-layered mappings, which we use subsequently to build multi-layered hierarchies in compositional manner. Firstly, we analyze the transition from a concrete state to an abstract state without considering subsequent layers of abstractions. This focused approach clarifies and classifies foundational abstraction methods, forming a basis for more complex, multi-layered abstractions. Restricting our examination to one-layered mappings, we categorize abstraction processes based on structural characteristics and mapping nature. These include Direct Abstraction (each concrete state maps to a unique abstract state), Divergent Abstraction (a single concrete state maps to multiple abstract states), and Convergent Abstraction (multiple concrete states aggregate into a single abstract state). Each type offers insights into how information and characteristics from the concrete domain are represented, reduced, or combined in the abstract domain. By focusing on these initial mappings, we establish a foundational taxonomy, facilitating the systematic exploration of abstraction strategies and laying the groundwork for investigating multi-layered hierarchical abstractions.

\subsection{Direct Abstraction (One-to-One)}
In our taxonomy, the first category, direct abstraction, entails a one-to-one mapping between each concrete state and its abstract counterpart. Holtzen et al.’s work on probabilistic program abstraction exemplifies this approach, using direct abstraction specifically in the context of probabilistic programs \cite{holtzen2017probabilistic}. Building on this example and retaining the same measure-theoretic approach, we generalize the framework to encompass a wider range of one-to-one mappings that extend beyond program-specific applications, thereby broadening the applicability of probabilistic abstraction, as outlined schematically below.
$$
\centering
\begin{array}{cc}
(\Omega_{a}, \Sigma_{a}, \mu_{a}) \\
\bigg{\uparrow^{[\![\mathcal{A}]\!]}}  \\
(\Omega_{c}, \Sigma_{c}, \mu_{c})
\end{array}
$$

\textbf{Definition 1 (Direct Abstraction)}: Let \((\Omega_{c}, \Sigma_{c}, \mu_{c})\) be a concrete probability space and \((\Omega_{a}, \Sigma_{a}, \mu_{a})\) an abstract probability space. Direct abstraction is facilitated by a bijective measurable function \(\mathcal{A}: \Omega_{c} \rightarrow \Omega_{a}\), ensuring a one-to-one correspondence between elements of \(\Omega_{c}\) and \(\Omega_{a}\).

Formally, \(\mathcal{A}\) is a measurable function such that for any measurable set \(B \in \Sigma_{a}\), the pre-image \(\mathcal{A}^{-1}(B)\) is measurable in \(\Sigma_{c}\), i.e., \(\forall B \in \Sigma_{a}, \mathcal{A}^{-1}(B) \in \Sigma_{c}\). The abstract measure \(\mu_{a}\) is the pushforward of the concrete measure \(\mu_{c}\) via \(\mathcal{A}\), defined by \(\mu_{a}(B) = \mu_{c}(\mathcal{A}^{-1}(B))\) for all \(B \in \Sigma_{a}\), ensuring the preservation of probabilistic information through the abstraction process.

\subsection{Divergent Abstraction (One-to-Many)}
Divergent abstraction allows a single set of low-level evidence to be abstracted into multiple high-level theories or concepts, as illustrated below:  
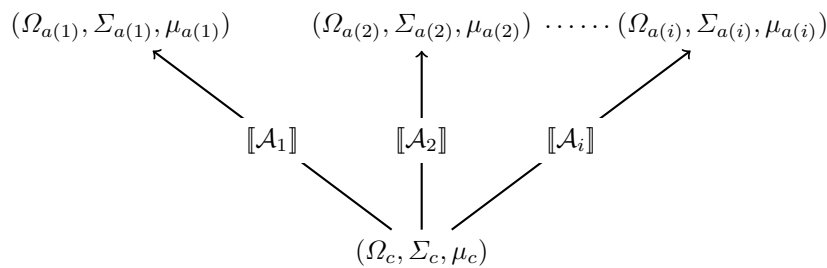
\[
\centering
\begin{tikzpicture}
    % Central Node at Bottom
    \node (C) at (0,-3) {$(\Omega_{c}, \Sigma_{c}, \mu_{c})$};
    
    % Abstract States at Top
    \node (A1) at (-4,0) {$(\Omega_{a(1)}, \Sigma_{a(1)}, \mu_{a(1)})$};
    \node (A2) at (0,0) {$(\Omega_{a(2)}, \Sigma_{a(2)}, \mu_{a(2)})$};
    \node at (2.1,0) {$\dots \dots$}; % For indicating continuation
    \node (Ai) at (4,0) {$(\Omega_{a(i)}, \Sigma_{a(i)}, \mu_{a(i)})$};
    
    % Arrows from Abstract States to Central Node
    \draw[<-, thick] (A1) -- (C) node [midway, fill=white] {$[\![\mathcal{A}_{1}]\!]$};
    \draw[<-, thick] (A2) -- (C) node [midway, fill=white] {$[\![\mathcal{A}_{2}]\!]$};
    \draw[<-, thick] (Ai) -- (C) node [midway, fill=white] {$[\![\mathcal{A}_{i}]\!]$};
\end{tikzpicture}
\]

\textbf{Definition 2 (Divergent Abstraction)}: 
Given a concrete probability space \((\Omega_{c}, \Sigma_{c}, \mu_{c})\), divergent abstraction maps this space to a collection of abstract probability spaces \(\{(\Omega_{a(i)}, \Sigma_{a(i)}, \mu_{a(i)})\}_{i \in I}\), where \(I\) is an index set, via measurable functions \(\{\mathcal{A}_i: \Omega_{c} \rightarrow \Omega_{a(i)}\}_{i \in I}\). Each \(\mathcal{A}_i\) is measurable, ensuring that for every \(i \in I\) and \(A_i \in \Sigma_{a(i)}\), \(\mathcal{A}_i^{-1}(A_i) \in \Sigma_{c}\). The unified abstract sigma-algebra \(\Sigma^{'}\) is defined as \(\sigma\left(\bigcup_{i \in I} \mathcal{A}_i(\Sigma_{c})\right)\), the smallest sigma-algebra containing the union of the images of \(\Sigma_{c}\) under all \(\mathcal{A}_i\). A unified abstract measure \(\mu^{'}\) on \(\Sigma^{'}\) is constructed such that for any \(E \in \Sigma^{'}\), \(\mu^{'}(E)\) integrates or aggregates \(\mu_{a(i)}(E \cap \Omega_{a(i)})\) for all \(i \in I\), normalized to ensure \(\mu^{'}\) is a probability measure.

\subsection{Convergent Abstraction (Many-to-One)}
Conversely, in convergent abstraction, multiple sets of low-level evidence are integrated into a single high-level theory, allowing for a comprehensive synthesis of low-level evidence through the abstraction process as illustrated below: 

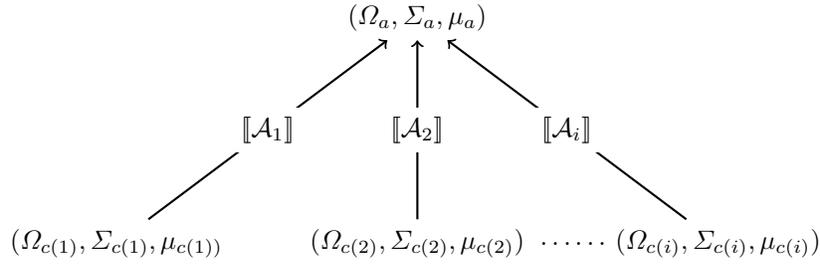
\[
\centering
\begin{tikzpicture}
    % Central Node at Top
    \node (C) at (0,3) {$(\Omega_{a}, \Sigma_{a}, \mu_{a})$};
    
    % Abstract States at Bottom
    \node (A1) at (-4,0) {$(\Omega_{c(1)}, \Sigma_{c(1)}, \mu_{c(1))}$};
    \node (A2) at (0,0) {$(\Omega_{c(2)}, \Sigma_{c(2)}, \mu_{c(2)})$};
    \node at (2.1,0) {$\dots \dots$}; % For indicating continuation
    \node (Ai) at (4,0) {$(\Omega_{c(i)}, \Sigma_{c(i)}, \mu_{c(i)})$};
    
    % Arrows from Abstract States to Central Node
    \draw[->, thick] (A1) -- (C) node [midway, fill=white] {$[\![\mathcal{A}_{1}]\!]$};
    \draw[->, thick] (A2) -- (C) node [midway, fill=white] {$[\![\mathcal{A}_{2}]\!]$};
    \draw[->, thick] (Ai) -- (C) node [midway, fill=white] {$[\![\mathcal{A}_{i}]\!]$};
\end{tikzpicture}
\]

\textbf{Definition 3 (Convergent Abstraction)}: Let \(\{(\Omega_{c(i)}, \Sigma_{c(i)}, \mu_{c(i)})\}_{i \in I}\) be a collection of concrete probability spaces and \((\Omega_{a}, \Sigma_{a}, \mu_{a})\) an abstract probability space. Convergent abstraction is achieved through measurable functions \(\{\mathcal{A}_i: \Omega_{c(i)} \rightarrow \Omega_{a}\}_{i \in I}\), which map elements from \(\Omega_{c(i)}\) to \(\Omega_{a}\), thereby aggregating disparate concrete spaces into a single abstract space. The formal characterization includes two main criteria: first, each \(\mathcal{A}_i\) must be measurable, ensuring that for every \(B \in \Sigma_{a}\), the pre-image \(\mathcal{A}_i^{-1}(B)\) is measurable in \(\Sigma_{c(i)}\); formally, \(\forall i \in I, \forall B \in \Sigma_{a}, \mathcal{A}_i^{-1}(B) \in \Sigma_{c(i)}\). Second, the abstract measure \(\mu_{a}\) integrates the measures from all concrete spaces, defined for any \(B \in \Sigma_{a}\) as \(\mu_{a}(B) = \sum_{i \in I} \mu_{c(i)}(\mathcal{A}_i^{-1}(B))\), with appropriate normalization to ensure \(\mu_{a}\) remains a probability measure, thus coherently consolidating probabilistic information from multiple concrete spaces into the abstract space.

\section{Advancing into Hierarchical Probabilistic Abstraction}
Using the foundational types of abstraction (direct, convergent, and divergent) various hierarchical taxonomies can be constructed by combining these single-layered abstractions into structured, tree-like models. Layering direct abstractions sequentially creates a multi-layered hierarchy known as sequential abstraction, where each level builds directly on the previous one, enabling a systematic progression from low-level details to high-level concepts.

Beyond simple sequential structures, hybrid hierarchical probabilistic abstractions introduce more complex configurations. For instance, a model beginning with a divergent layer followed by sequential direct abstractions is termed a divergent sequential abstraction. A divergent sequential abstraction enables multiple sequential abstractions on the same foundational data, whereas a convergent sequential abstraction starts with different foundational spaces, applies sequential abstractions to each, and unifies them at higher levels.

This structured approach enables the construction of hierarchical models that maintain consistency and interpretability across abstraction layers, supporting diverse modeling needs. In certain cases, divergent abstractions may even simplify into a sequential form when no further divergent paths exist, underscoring the importance of understanding structural relationships between abstraction types. By grounding higher-level abstractions in low-level details, we ensure that the entire hierarchy retains coherence and accuracy across layers.

In our approach to representing hierarchical abstract representations, we categorize into two types of Hierarchical Probabilistic Abstraction Models (HPAMs) based on the complexities they capture:
\begin{enumerate}
 \item  HPAM-DAG (Directed Acyclic Graphs) \footnote{The decision to employ DAG-based models is motivated by their ability to enable a clear, systematic transition from concrete to abstract representations in a unidirectional manner, facilitating straightforward analysis and interpretation of probabilistic relationships.}: Uses a tree-like structure for systems with clear, sequential progressions and no cycles. This type is ideal for unidirectional hierarchical relationships, simplifying analysis.
 \item  HPAM-CD (Cyclical and Dynamic): Captures systems with feedback loops, cycles, and dynamic interactions, reflecting real-world complexity. Due to these complexities, this type requires more formal and mathematical treatment and is reserved for future work.
\end{enumerate}

For now, we focus exclusively on HPAM-DAGs to simplify and analyze complex probabilistic systems, ensuring coherent, unidirectional mapping and maintaining clear hierarchical relationships.\\

\textbf{Definition 4 (HPAM-DAG):}
A Hierarchical Probabilistic Abstraction, denoted as $\mathcal{H}$, is defined as a triple $\left( \mathcal{V}, \mathcal{E}, \{\mathcal{P}_v\}_{v \in \mathcal{V}} \right)$ where:

\begin{itemize}
    \item $\mathcal{V}$ is a set of vertices in a Directed Acyclic Graph (DAG), each corresponding to a distinct probabilistic space.
    \item $\mathcal{E} \subseteq \mathcal{V} \times \mathcal{V}$ is a set of directed edges in the DAG, each representing an abstraction mapping between probabilistic spaces.
    \item $\{\mathcal{P}_v\}_{v \in \mathcal{V}}$ is a family of probabilistic spaces associated with vertices in $\mathcal{V}$. For each vertex $v$, the probabilistic space $\mathcal{P}_v$ is a tuple $(\Omega_v, \Sigma_v, P_v)$, where:
    \begin{itemize}
        \item $\Omega_v$ is the sample space for vertex $v$, representing all possible outcomes.
        \item $\Sigma_v$ is a $\sigma$-algebra over $\Omega_v$, defining the set of events for which probabilities are assigned.
        \item $P_v: \Sigma_v \rightarrow [0, 1]$ is a probability measure that assigns probabilities to events in $\Sigma_v$.
    \end{itemize}
\end{itemize}

For every directed edge $(v_i, v_j) \in \mathcal{E}$ within the DAG, there exists an abstraction mapping $\mathcal{A}_{ij}: \Omega_{v_i} \rightarrow \Omega_{v_j}$ that facilitates a transformation between probabilistic spaces $\mathcal{P}_{v_i}$ and $\mathcal{P}_{v_j}$. This mapping satisfies the condition:
\begin{equation}
    P_{v_j}(A_{v_j}) = P_{v_i}(\mathcal{A}_{ij}^{-1}(A_{v_j})),
\end{equation}
for all $A_{v_j} \in \Sigma_{v_j}$, where $\mathcal{A}_{ij}^{-1}(A_{v_j})$ is the pre-image of $A_{v_j}$ under $\mathcal{A}_{ij}$, ensuring the preservation of probabilistic measures through the abstraction process.

\subsection{Boundary of Abstraction}
 
 Establishing a clear boundary for abstraction in probabilistic models is essential, and the Highest Possible Abstraction (HPoA) serves to mark this boundary. Beyond the HPoA, further abstraction risks obscuring critical probabilistic or relational details, diminishing the model's interpretability and analytical rigor. The HPoA thus defines the upper limit of meaningful abstraction, ensuring that each layer within the abstraction hierarchy contributes substantively to the model's integrity. This approach preserves essential information at every level, supporting the construction of hierarchical models that are both robust and interpretable.\\

\textbf{Definition 5 (HPoA).}
Let's consider a HPAM-DAG denoted as $\mathcal{H} = \left( \mathcal{V}, \mathcal{E}, \{\mathcal{P}_v\}_{v \in \mathcal{V}} \right)$. Within this framework, HPoA is formalized as a specific probabilistic space $(\Omega_{HPoA}, \Sigma_{HPoA}, P_{HPoA})$ that meets the following rigorously defined criteria:

\begin{enumerate}
    \item \textbf{Preservation of Probabilistic Integrity}: For every set $A_{HPoA} \in \Sigma_{HPoA}$, it holds that $P_{HPoA}(A_{HPoA})$ is congruent with the probabilistic measure assigned to its corresponding set in any precursor probabilistic space within $\mathcal{H}$. Formally, this means:
    \[ \begin{array}{c} 
    \forall A_{HPoA} \in \Sigma_{HPoA}, \exists A_{\mathcal{P}_v} \in \Sigma_{\mathcal{P}_v} \text{ for } \mathcal{P}_v \in \{\mathcal{P}_v\}_{v \in \mathcal{V}}: \\ P_{HPoA}(A_{HPoA}) = P_{\mathcal{P}_v}(\mathcal{A}_{v \rightarrow HPoA}^{-1}(A_{HPoA})),
    \end{array}
    \]
    where $\mathcal{A}_{v \rightarrow HPoA}$ denotes the abstraction mapping from $\Omega_v$ to $\Omega_{HPoA}$.
    
    \item \textbf{Maximal Generalization}: $(\Omega_{HPoA}, \Sigma_{HPoA}, P_{HPoA})$ achieves the highest degree of abstraction permissible under the constraints of probabilistic integrity. No further abstraction $(\Omega', \Sigma', P')$ can be derived from HPoA without a resultant loss in the preservation of essential probabilistic relationships. This is expressed as:
    $ \not\exists (\Omega', \Sigma', P') : \Sigma_{HPoA} \subsetneq \Sigma' \land \forall A' \in \Sigma', P'(A') \neq P_{HPoA}(\mathcal{A}_{HPoA \rightarrow '}^{-1}(A')).$
\end{enumerate}

In essence, the HPoA, $(\Omega_{HPoA}, \Sigma_{HPoA}, P_{HPoA})$, encapsulates the terminus of the abstraction process within $\mathcal{H}$, marking the juncture beyond which no additional abstraction can be achieved without compromising the model’s fundamental probabilistic structure.\\

\textbf{Proposition 1 (Existence of the HPoA in HPAM-DAG).}
Given a HPAM-DAG $\mathcal{H} = (\mathcal{V}, \mathcal{E}, \{\mathcal{P}_v\}_{v \in \mathcal{V}})$, there exists at least one HPoA, $(\Omega_{HPoA}$ $,\Sigma_{HPoA},P_{HPoA})$, such that for any probabilistic space $(\Omega', \Sigma', P')$ that can be derived from $(\Omega_{HPoA}, \Sigma_{HPoA}, P_{HPoA})$ via an abstraction mapping, it results in the loss of at least one essential probabilistic relationship.

\section{HPAM-DAG Types}
Different types of HPAM-DAGs emerge as we combine foundational abstraction types in various ways. Each variant of HPAM-DAG has unique properties based on the specific combination and configuration of these foundational mappings. By arranging these mappings into multilayered structures, we can create HPAM-DAGs that represent increasing levels of complexity within probabilistic systems.

In this section, we outline two types of HPAM-DAGs: sequential abstraction and a more complex hybrid variant. While these examples provide a foundation, it is important to note that many other types of HPAM-DAGs can be formally defined by varying combinations of foundational abstraction types. These additional configurations offer potential avenues for exploration in future works.

\subsection{Sequential Abstraction}
Sequential abstraction represents the simplest form of HPAM, as it layers direct abstractions in a straightforward, linear progression from low-level to high-level concepts, without incorporating branching or merging complexities.\\

\textbf{Definition 6 (Sequential Abstractions):} Let $(\Omega_0, \Sigma_0, P_0)$ represent the foundational probability space. Sequential abstraction is characterized by a finite series of probability spaces $\{(\Omega_i, \Sigma_i, P_i)\}_{i=0}^n$, where each space $(\Omega_{i+1},$ $\Sigma_{i+1}, P_{i+1})$ is derived from $(\Omega_i, \Sigma_i, P_i)$ via an abstraction operation $\mathcal{A}_i$. The culmination of this process, the HPoA, $(\Omega_{HPoA}, \Sigma_{HPoA},P_{HPoA})$, is distinguished as the endpoint of the sequence. For each $i$, ranging from $1$ to $n$, this is formally represented as:
\[
\mathcal{A}_i: (\Omega_{i-1}, \Sigma_{i-1}, P_{i-1}) \rightarrow (\Omega_i, \Sigma_i, P_i),
\]
under the stipulations that: (1) for every measurable set \( A \in \Sigma_{i-1} \), the condition \( P_{i-1}(A) = P_i(\mathcal{A}_i(A)) \) holds true (Preservation of Probability Mass), and (2) there is no \(\sigma\)-algebra smaller than \(\Sigma_i\), denoted as \(\Sigma' \subsetneq \Sigma_i\), for which the preservation of probability mass criteria remains valid (Minimality).\\ 

\textbf{Proposition 2 (Uniqueness of HPoA in Sequential Abstractions):} Given a concrete probability space $(\Omega_0, \Sigma_0, P_0)$, the HPoA,  $(\Omega_{HPoA}, \Sigma_{HPoA},$ $ P_{HPoA})$, achieved through a sequential abstraction process is unique. This implies that for any two HPoAs,  $(\Omega_{HPoA}, \Sigma_{HPoA}, P_{HPoA})$ and $(\Omega'_{HPoA}, \Sigma'_{HPoA},$ $ P'_{HPoA})$ derived from $(\Omega_0, \Sigma_0, P_0)$ via any sequence of abstraction operations, it must hold that $\Sigma_{HPoA} = \Sigma'_{HPoA}$.\\

\textbf{Proposition 3 (Existence of Intermediate States):}
Given a concrete probability space $(\Omega_0, \Sigma_0, P_0)$ and a HPoA, $(\Omega_{HPoA}, \Sigma_{HPoA}, P_{HPoA})$, obtained through a sequential abstraction process, for any intermediate abstraction $(\Omega_i, \Sigma_i$ $, P_i)$ where $0 < i < n$ and $(\Omega_i, \Sigma_i, P_i) \neq (\Omega_{HPoA}, \Sigma_{HPoA}, P_{HPoA})$, there exists at least one intermediate state $(\Omega_{int}, \Sigma_{int}, P_{int})$ that facilitates the decomposition of the abstraction sequence into:
\[
[\![\mathcal{A}_{\text{pre}}]\!]: \Omega_0 \rightarrow \Omega_{int} \quad \text{and} \quad [\![\mathcal{A}_{\text{post}}]\!]: \Omega_{int} \rightarrow \Omega_{HPoA}.
\]

\textbf{Property 1 (Comprehensibility):}
Given the existence of an intermediate abstraction state \((\Omega_{int}, \Sigma_{int}, P_{int})\), there exists a structured pathway, \([\![\mathcal{A}_{\text{pre}}]\!]: \Omega_0 \rightarrow \Omega_{int}\) and \([\![\mathcal{A}_{\text{post}}]\!]: \Omega_{int} \rightarrow \Omega_{HPoA}\), that enhances the comprehensibility of the abstraction process. This is formalized as the ability to sequentially deconstruct the abstraction hierarchy into comprehensible steps, each represented by measurable transformations preserving essential probabilistic structures.

Inspired by Ai et al. findings on sequential learning tasks, our abstraction process is designed with intermediate stages to enhance comprehensibility \cite{muggleton_sequential}. Ai et al. noted that structured and sequential presentation of concepts improves human understanding. By preserving intermediate abstraction states, we ensure each step remains comprehensible and accessible.\\

\textbf{Property 2 (Tractability):}
The identification and utilization of the intermediate abstraction state $(\Omega_{int}, \Sigma_{int}, P_{int})$ within the sequential abstraction process confirm the modular nature of the abstraction and significantly improve the tractability of conducting analyses across the hierarchy. This modularity allows for targeted adjustments and refinements at specific levels of abstraction, effectively navigating and understanding probabilistic relationships at each hierarchical level.

Ensuring tractability is vital for the practical application of abstraction methodologies. Results from Holtzen et al. demonstrate that maintaining computational manageability at each step allows for efficient analysis and reasoning, making the overall process more effective \cite{holtzen2018sound,holtzen2017probabilistic}.

\subsection{Hybrid HPAMs}

Hybrid HPAM-DAGs amalgamate sequential, divergent, and convergent abstraction methodologies within hierarchical probabilistic modeling to adeptly capture the complexities of intricate systems. Defined over a base probability space \((\Omega_0, \Sigma_0, P_0)\), Hybrid HPAM-DAG employs a dynamic mix of abstraction operations that either progress linearly, branch out divergently, and/or converge from various sequences into a unified model. This integration allows for a nuanced exploration and understanding of the system's probabilistic behaviors from multiple perspectives, ensuring the model comprehensively reflects the system's dynamics and dependencies. By leveraging the strengths of each abstraction method, Hybrid HPAM-DAG offers a versatile framework that maintains the system's probabilistic integrity while navigating through its complex landscape, making it a powerful tool for modeling sophisticated systems with high fidelity.

To exemplify, consider a hybrid model that combines direct, sequential, divergent, and convergent abstraction processes to model the dynamic nature of complex systems like Alzheimer's disease. It begins with a foundational concrete space \((\Omega_0, \Sigma_0, P_0)\), where \(\Omega_0\) represents the population at risk for Alzheimer's disease, \(\Sigma_0\) includes measurable events indicating risk factors for Alzheimer's, and \(P_0\) quantifies the initial distribution of these risk factors. For more details, refer to the appendix section for a pseudocode algorithm.   

\subsubsection{Sequential Abstraction:} On the first layer, the abstraction transitions the model from broad risk factors to specific biological markers indicative of Alzheimer's:
\[
\mathcal{A}_1: (\Omega_0, \Sigma_0, P_0) \rightarrow (\Omega_1, \Sigma_1, P_1)
\]
where \((\Omega_1, \Sigma_1, P_1)\) focuses on biological markers such as amyloid-beta levels and tau protein tangles. The model employs sequential abstractions to refine the understanding of how these biological markers influence disease progression:
\[
\mathcal{A}_2: (\Omega_1, \Sigma_1, P_1) \rightarrow (\Omega_2, \Sigma_2, P_2)
\]
where \((\Omega_2, \Sigma_2, P_2)\) might include cognitive decline metrics and early symptoms of Alzheimer's disease.

\subsubsection{Divergent Abstraction:} From this point, the model branches into distinct intervention pathways:
\[
\mathcal{A}_{3,i}: (\Omega_2, \Sigma_2, P_2) \rightarrow (\Omega_{3,i}, \Sigma_{3,i}, P_{3,i}), \quad i \in \{1, 2\}
\]
Each path \(i\) represents a different intervention strategy, such as cognitive therapy or medication, with outcomes specific to those treatments. For example:
\begin{enumerate}
    \item \(\mathcal{A}_{3,1}\) could represent cognitive therapy, focusing on improving cognitive functions through exercises and mental activities.
    \item \(\mathcal{A}_{3,2}\) could represent medication, focusing on pharmacological treatments aimed at slowing the progression of the disease.
\end{enumerate}

\subsubsection{Convergent Abstraction:}The insights from these divergent paths are then synthesized into a unified model:
\[
\mathcal{A}_4: (\Omega_{3,1} \cup \Omega_{3,2}, \Sigma_{3,1} \cup \Sigma_{3,2}, P_{3,1} \oplus P_{3,2}) \rightarrow (\Omega_4, \Sigma_4, P_4)
\]
resulting in a comprehensive understanding of the efficacy and outcomes of various Alzheimer's disease management strategies.

This hybrid methodology can be applied in several contexts: integrating genetic, environmental, and clinical data to study Alzheimer's disease progression and treatment effectiveness in research; developing personalized treatment plans that consider multiple intervention strategies in clinical practice; and informing healthcare policies based on comprehensive models that incorporate various risk factors and treatment outcomes in policy making. The Hybrid HPAM-DAG framework effectively integrates sequential, divergent, and convergent abstraction processes, providing a structured approach to understanding and managing Alzheimer's disease. It progressively abstracts and refines data, offering valuable insights at each hierarchical level.

\section{Related Works}
Abstraction serves as a critical tool for simplifying and understanding the complexity inherent in drawing analogies, parsing through inherent relational information, employing reasoning by formally representing uncertainty \cite{saitta2013abstraction,reasoning-about-uncertainty,analogyAndAbstraction}. The work of Saitta and Zucker, for instance, provides a comprehensive exploration of abstraction, from its fundamental role in planning and solving constraint satisfaction problems to its implementation in the coordination of multi-agent systems \cite{saitta2013abstraction}. Moreover, abstraction plays a crucial role in domains such as program analysis and verification, probabilistic games, labeled transition systems, probabilistic programs, complex systems, and inductive logic programming, enhancing the innovative problem-solving capabilities and methodological developments within each field \cite{banihashemi2017abstraction,cropper2016learning,gameAbstraction,halpern2004representation,cousot2012probabilistic,monniaux2001abstract,inductive-logic-programming:-theory}. While some research has concentrated on the specific challenges and opportunities presented by probabilistic settings, others have offered theoretical insights on ideal specifications and the dynamics of mappings within abstractions more generally \cite{mciver2005abstraction,Dehnert2014,probabilistic-graphical-models:-principles}. Through these diverse treatments, abstraction emerges not only as a key concept for theoretical exploration but also as a practical tool for accommodating complex problem-solving methodologies across a spectrum of scientific inquiries \cite{machine-learning:-a-probabilistic-perspective}.

Numerous fields have developed sophisticated frameworks to formalize abstraction, providing a mathematical foundation for this complex concept and offering structured analyses of the abstraction process \cite{giunchiglia1992theory,banihashemi2017abstraction}. Milner’s work on the semantics of concurrent processes has significantly influenced our exploration of algebraic formalisms for probabilistic abstraction \cite{milner1990operational}. Similarly, Hennessy and Milner’s algebraic laws, which introduce observational congruence to study nondeterministic and concurrent programs, offer valuable insights into comparing different levels of abstractions within probabilistic abstraction hierarchies \cite{milnerAlgebraiclaws}. Further, studies on probabilistic bisimulation and cocongruence for probabilistic systems could inspire novel approaches in this area as exemplified by the work by Hofmann and Belle on abstracting noisy robot programs \cite{danos2006bisimulation,panangaden2015probabilistic,Hofmann_Belle,bisimAndAbstraction}. The algebraic tools as discussed, developed for defining and verifying concurrent program properties, hold promise for extending into designing and verifying probabilistic abstraction hierarchies. However, practical application of these theories has been mixed—some have been successfully integrated into real-world applications, while others face challenges in bridging theoretical insights with practical implementation \cite{giunchiglia1992theory,mciver2005abstraction,clarke2000counterexample}.

In Belle's work, for instance, a novel semantic framework for analyzing and abstracting probabilistic models is unveiled, significantly contributing to the discourse on abstraction \cite{Abstraction_Belle}. This research builds upon and extends the foundational work of Banihashemi et al., which investigated isomorphism and thorough abstractions in the context of situation calculus agent programs, primarily from a non-probabilistic perspective \cite{banihashemi2017abstraction}. Belle aims to adapt these foundational concepts to probabilistic settings, such as in probabilistic relational models (PRMs), through the introduction of unweighted abstractions that are compatible with categorical settings in terms of satisfaction and entailment \cite{Abstraction_Belle,banihashemi2017abstraction}. This foundation facilitates the advancement of weighted abstractions and the integration of evidence, while also examining the linkage between these abstractions and stochastic models, notably through the lens of weak exact abstractions and the automation of abstraction generation.

Moreover, Belle also highlights the potential of incorporating methodologies from Giunchiglia and Walsh's research on logical theory operations into the probabilistic framework, necessitating some adjustments \cite{giunchiglia1992theory,Abstraction_Belle}. Albeit with some limitations in the unweighted and weighted approaches outlined, Belle provides detailed discussions on Banihashemi et al.'s approach of abstraction in knowledge representation and automated planning, especially hierarchical planning. Belle's work opens new avenues for both theoretical exploration and practical application within probabilistic modeling and planning \cite{giunchiglia1992theory,banihashemi2017abstraction,Abstraction_Belle,HierarchialProb_Abs_Refinement}. We start our inquiry by following up on the call for newer approaches in representing abstraction hierarchies as Belle and Banihashemi et al. outline \cite{banihashemi2017abstraction,Abstraction_Belle}.

Further, Holtzen et al. explores the concept of abstraction within probabilistic programming, specifically investigating how to abstract a probability distribution defined by such a program \cite{holtzen2018sound,holtzen2017probabilistic}. The study identifies and automatically generates abstractions that semantically represent conditional independence assumptions, thereby aiding in the design of algorithms for simplifying inference processes. This is achieved by adapting techniques from program verification to analyze deterministic code, applying a refined form of predicate abstraction to create abstract probabilistic programs that preserve the essential characteristics of the original programs \cite{holtzen2018sound} . A key contribution of their work is the introduction of the concept of distributional soundness, which ensures the consistency of probability distributions between abstract and concrete programs. Holtzen et al. outlines a theory and methodology for developing distributionally sound abstractions across a broad range of probabilistic programs, leveraging complex independence structures to streamline the original program \cite{holtzen2018sound}. Through this approach, Holtzen et al. illustrates the practical advantages of abstraction, such as enhanced efficiency in inference procedures, across various statistical models formulated as probabilistic programs \cite{holtzen2018sound,holtzen2017probabilistic}.

\section{Discussion}
  
Our approach seeks to unify diverse theoretical perspectives while advancing the understanding and practical application of abstraction, bridging the gap between theoretical foundations and real-world utility in hierarchical representations. Despite its strengths, our framework has several limitations. The integration of multiple abstraction layers and the hybrid nature of our framework can lead to increased computational complexity, potentially making it less scalable for very large datasets or highly complex systems. While our approach aims to enhance interpretability by structuring abstractions hierarchically, there is a trade-off between the level of detail retained and the simplicity of the abstracted models, making it challenging to balance these aspects. Furthermore, our framework, though robust in the contexts we've explored, may face challenges when generalized to entirely new domains or types of data that were not considered during its development.

The effectiveness of our framework heavily relies on the quality and completeness of the initial data. Poor quality data can lead to inaccurate abstractions and potentially flawed conclusions. Additionally, the hierarchical nature and the need for intermediate abstractions require significant computational resources, which may not be readily available in all settings. The current use of Directed Acyclic Graphs (DAGs) for representation, while useful for capturing dependencies and causal relationships, can be limiting in expressing more complex, multi-layered abstractions. A more rigorous and holistic framework is necessary to represent abstraction hierarchies effectively, ensuring that multi-layered representations are comprehensively understood and utilized.

This work introduces the concepts of hierarchical probabilistic abstraction and a specific hybrid approach, yet many other taxonomies and abstraction methods remain unexplored and warrant further study to fully understand abstraction frameworks. Recognizing the importance of hierarchical representations for interpreting multi-layered abstractions, future work will provide a formal treatment of this framework by developing a rigorous mathematical foundation and delivering a comprehensive analysis of the Hybrid HPAM-DAG approach. We also intend to address current limitations by refining computational strategies, investigating additional taxonomies of abstraction, and conducting extensive testing across diverse domains and applied settings. These efforts will enhance our approach, expand its applicability, and ultimately lead to a more robust and holistic representation of abstraction hierarchies.

\section{Conclusion}  
In this paper, we introduce a conceptual framework for hierarchical probabilistic abstraction that extends a measure-theoretic foundation to address challenges in modeling complex systems. By structuring the abstraction process into layered mappings, this framework supports modular problem-solving and allows for both detailed layer-specific analysis and a cohesive, system-wide understanding. This dual-level approach enhances interpretability and computational tractability by bridging high-level conceptual insights with low-level perceptual data. The modular structure ensures that each layer of the abstraction hierarchy can be independently developed, analyzed, and comprehended, fostering the construction of interpretable and tractable models. Additionally, this hierarchical framework provides flexibility in integrating multiple abstraction levels, facilitating the exploration of complex phenomena with both depth and breadth.

\bibliographystyle{splncs04}
\bibliography{references}

\newpage
\appendix

\section{Hybrid HPAM for Alzheimer’s Disease Management}
In this section, we provide a high-level pseudocode algorithm to illustrate the application of HPAM-DAGs in Alzheimer's disease management. The algorithm demonstrates how direct, divergent, and convergent abstractions can model the relationships among risk factors, biological markers, treatment pathways, and patient outcomes.

\subsection*{Pseudocode Algorithm}

\begin{itemize}
\scriptsize
    \item[1:] Initialize DAG $G$ with initial probability space $(\Omega_0, \Sigma_0, P_0)$.
    \item[2:] Initialize $(\text{Final\_Model}, \text{HPoA}) = \{\}$, $\text{HPoA} = \text{undefined}$.
    \item[3:] \textbf{while} True \textbf{do}
    \begin{itemize}
        \item[4:] \textbf{Step 1: Apply Direct Abstraction} to map low-level risk factors to biological markers.
        \item[5:] \textbf{for each} risk\_factor in $(\Omega_0, \Sigma_0, P_0)$ \textbf{do}
        \begin{itemize}
            \item[6:] Define Direct Abstraction:
            \[
            (\Omega_1, \Sigma_1, P_1) = \mathcal{A}_\text{Direct}((\Omega_0, \Sigma_0, P_0))
            \]
            \item[7:] Add Node $G: (\Omega_1, \Sigma_1, P_1)$.
            \item[8:] Add Edge $G: (\Omega_0, \Sigma_0, P_0) \rightarrow (\Omega_1, \Sigma_1, P_1)$.
        \end{itemize}
        \item[9:] \textbf{end for}

        \item[10:] \textbf{Step 2: Apply Divergent Abstraction} to generate multiple treatment pathways.
        \item[11:] \textbf{for each} $(\Omega_1, \Sigma_1, P_1)$ in Bio\_Markers \textbf{do}
        \begin{itemize}
            \item[12:] Define Divergent Abstraction:
            \[
            (\Omega_{2_i}, \Sigma_{2_i}, P_{2_i}) = \mathcal{A}_\text{Divergent}((\Omega_1, \Sigma_1, P_1))
            \]
            \item[13:] Add Node $G: (\Omega_{2_i}, \Sigma_{2_i}, P_{2_i})$.
            \item[14:] Add Edge $G: (\Omega_1, \Sigma_1, P_1) \rightarrow (\Omega_{2_i}, \Sigma_{2_i}, P_{2_i})$.
        \end{itemize}
        \item[15:] \textbf{end for}

        \item[16:] \textbf{Step 3: Apply Convergent Abstraction} to unify treatment outcomes.
        \item[17:] Combine using Convergent Abstraction:
        \[
        (\Omega_3, \Sigma_3, P_3) = \mathcal{A}_\text{Convergent}(\{(\Omega_{2_1}, \Sigma_{2_1}, P_{2_1}), \dots, (\Omega_{2_n}, \Sigma_{2_n}, P_{2_n})\})
        \]
        \item[18:] Add Node $G: (\Omega_3, \Sigma_3, P_3)$.
        \item[19:] Add Edge $G: \{(\Omega_{2_1}, \Sigma_{2_1}, P_{2_1}), \dots, (\Omega_{2_n}, \Sigma_{2_n}, P_{2_n})\} \rightarrow (\Omega_3, \Sigma_3, P_3)$.

        \item[20:] \textbf{Step 4: Calculate Highest Possible Abstraction (HPoA)}.
        \item[21:] Construct HPoA:
        \[
        (\Omega_\text{HPoA}, \Sigma_\text{HPoA}, P_\text{HPoA}) = \mathcal{A}_\text{HPoA}((\Omega_3, \Sigma_3, P_3))
        \]

        \item[22:] \textbf{Step 5: Validate if HPoA matches observed outcomes in Alzheimer’s patients}.
        \item[23:] \textbf{if} $P_\text{HPoA}$ matches observed outcomes $O$ \textbf{then}
        \begin{itemize}
            \item[24:] \quad \textbf{return} Success: Valid HPoA achieved for Alzheimer's Disease Management.
        \end{itemize}
        \item[25:] \textbf{else if} no further abstractions can be made
        \begin{itemize}
            \item[26:] \quad \textbf{return} Failure: HPoA cannot be constructed.
        \end{itemize}
        \item[27:] \textbf{else}
        \begin{itemize}
            \item[28:] \quad Perform interventions or update models and iterate.
            \item[29:] \quad Update DAG $G$ with new treatment data.
        \end{itemize}
        \item[30:] \textbf{end if}
    \end{itemize}
    \item[31:] \textbf{end while}
\end{itemize}

\end{document}